\documentclass[10pt,twocolumn]{article}
\usepackage{ifthen}
\usepackage[twoside,letterpaper,includeheadfoot,%
        layoutsize={8.125in,10.875in},%
                layouthoffset=0.1875in,%
                layoutvoffset=0.0625in,%
                left=43pt,%
                right=43pt,%
                top=43pt,
                bottom=43pt,%
                headheight=0pt,
                headsep=10pt,%
                footskip=25pt]{geometry}
\usepackage[footnotesize]{caption}
\usepackage{pdfpages}
\newboolean{arxiv}
\setboolean{arxiv}{true}                

\usepackage{graphicx,amsmath,amsfonts,amssymb}
\usepackage{hyperref}
\usepackage{apacite}
\usepackage{fancyhdr}

\usepackage{outlines}
\usepackage{apacite}
\usepackage{multirow}
\usepackage{tabularx}
\usepackage{booktabs}
\usepackage{array}

\usepackage{soul} 
\usepackage{color}
\usepackage{xcolor,colortbl}
\definecolor{orange}{HTML}{FCBF64}
\definecolor{skin}{HTML}{E79E97}

\usepackage{natbib}
\usepackage[explicit]{titlesec}
\titleformat{\section}
  {\large\bfseries}
  {\thesection.}
  {0.5em}
  {#1}
  []
\titleformat{name=\section,numberless}
  {\large\bfseries}
  {}
  {0em}
  {#1}
  []  
\titleformat{\subsection}[runin]
  {\bfseries}
  {\thesubsection.}
  {0.5em}
  {#1. }
  []
\titleformat{\subsubsection}[runin]
  {\small\bfseries\itshape}
  {\thesubsubsection.}
  {0.5em}
  {#1. }
  []    
\titleformat{\paragraph}[runin]
  {\small\bfseries}
  {}
  {0em}
  {#1} 
\titlespacing*{\section}{0pc}{3ex \@plus4pt \@minus3pt}{5pt}
\titlespacing*{\subsection}{0pc}{2.5ex \@plus3pt \@minus2pt}{2pt}
\titlespacing*{\subsubsection}{0pc}{2ex \@plus2.5pt \@minus1.5pt}{2pt}
\titlespacing*{\paragraph}{0pc}{1.5ex \@plus2pt \@minus1pt}{12pt}

\ifthenelse{\boolean{arxiv}}
{
  \title{\vspace{0cm} \LARGE People infer recursive visual concepts from just a few examples}
  \date{}
  \author{ \large Brenden M. Lake$^1$ and Steven T. Piantadosi$^2$\\
  \normalsize $^1$Department of Psychology and Center for Data Science, New York University \\
  \normalsize $^2$Department of Psychology, University of California Berkeley\\
  }
}
{
  \title{\LARGE People infer recursive visual concepts from just a few examples}
  \date{}
  \author{ 
  \large \textbf{Brenden M. Lake}$^*$\\  
  Department of Psychology and Center for Data Science \\
  New York University \\
  brenden@nyu.edu\\ 
  \and
  \large \textbf{Steven T. Piantadosi} \\  
  Department of Psychology \\
  University of California Berkeley \\
  stp@berkeley.edu \\
  }
}

\fancypagestyle{alim}{\fancyhf{}\fancyhead[L]{In press at \emph{Computational Brain \& Behavior}}}

\begin{document}
\maketitle
\ifthenelse{\boolean{arxiv}}
{
  \thispagestyle{alim}
}
{}

\begin{abstract}
Machine learning has made major advances in categorizing objects in images, yet the best algorithms miss important aspects of how people learn and think about categories.
People can learn richer concepts from fewer examples, including causal models that explain how members of a category are formed. 
Here, we explore the limits of this human ability to infer causal ``programs'' -- latent generating processes with nontrivial algorithmic properties -- from one, two, or three visual examples. 
People were asked to extrapolate the programs in several ways, for both classifying and generating new examples.
As a theory of these inductive abilities, we present a Bayesian program learning model that searches the space of programs for the best explanation of the observations. 
Although variable, people's judgments are broadly consistent with the model and inconsistent with several alternatives, including a pre-trained deep neural network for object recognition, indicating that people can learn and reason with rich algorithmic abstractions from sparse input data.
\end{abstract}

\section{Introduction}

\begin{figure}[!t]
\ifthenelse{\boolean{arxiv}}
{\centering\includegraphics[width=3.4in]{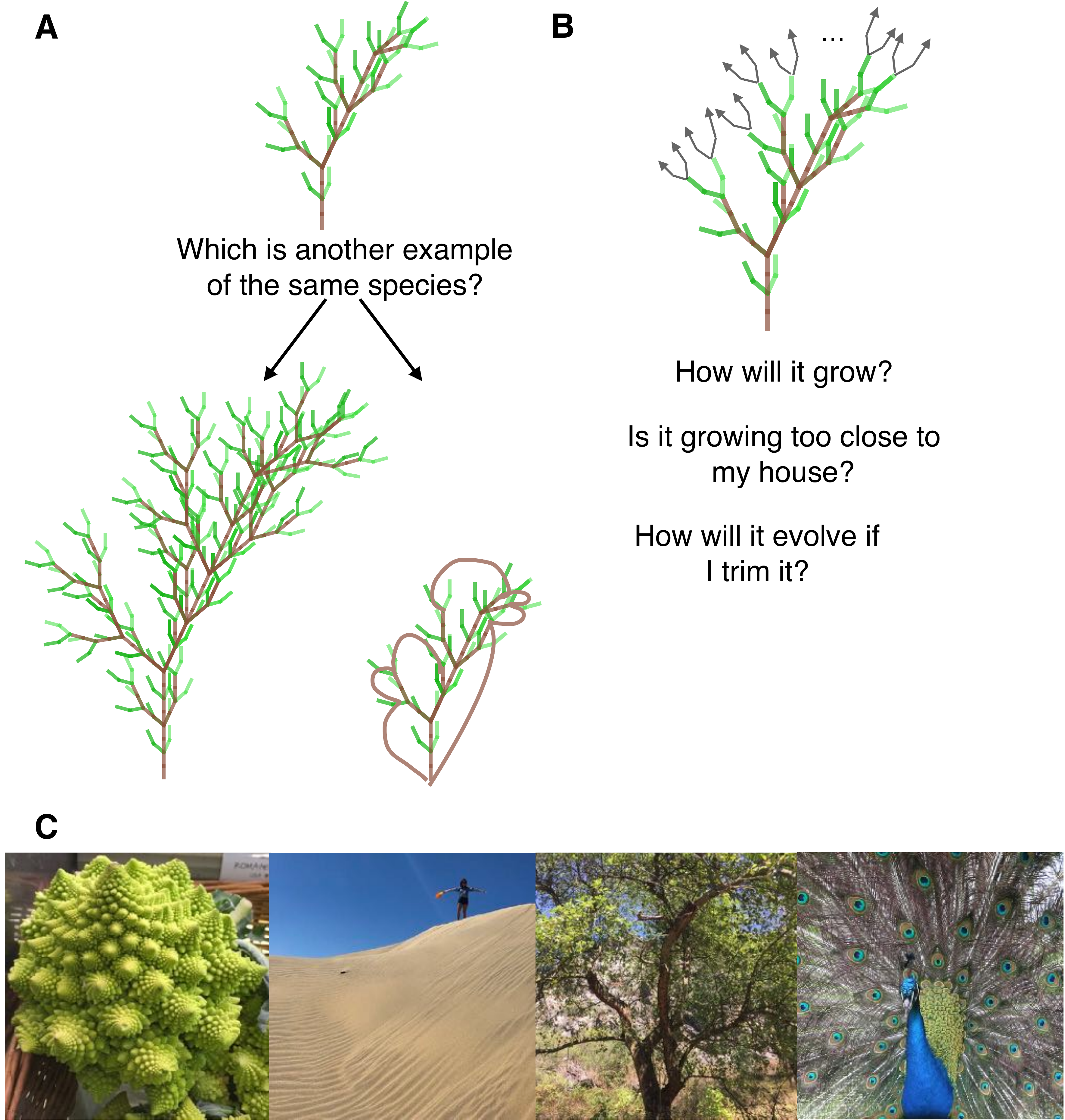}}
{\centering\includegraphics[width=4in]{figures/trees.pdf}}
\caption{Causal understanding influences everyday conceptual judgments in classification (A) and extrapolation (B). The top and left images of trees (A) have the same causal structure and were generated from the same simple program \protect\citep[L-system;][]{Prusinkiewicz1990}. However leading object recognition systems trained on natural images \protect\citep{Simonyan2014,He2015} understand little of that causal structure, perceiving the other two images as more similar (top and right) than the natural continuation (top and left; based on euclidean/cosine distance in the top hidden layer). (B) People also use their causal knowledge to make extrapolations, including predicting how trees grow. (C) In addition to trees, natural fractal concepts with rich causal structure include Romanesco, sand dunes, and peacock plumage.}
\label{fig:intro}
\end{figure}

Computer vision now approaches or exceeds human performance on certain large-scale object recognition tasks \citep{Krizhevsky2012,Russakovsky2014,Simonyan2014,He2015}, yet the best systems miss critical aspects of how people learn and think about categories \citep{Lake2016}. People learn richer and more abstract concepts than the best machines and, remarkably, they need fewer examples in order to do it. Children can make meaningful generalizations after seeing just one or a handful of ``pineapples'' or ``school buses'' \citep{smith-etal02,Bloom2000,Xu2007}, while the leading machine systems require far more examples. People can also use their concepts in more sophisticated ways than machines can -- not just for classification (Fig. \ref{fig:intro}A), but also for generation, explanation, and extrapolation (Fig. \ref{fig:intro}B). A central challenge is understanding how people learn such rich concepts from such limited amounts of experience.

An important limitation of contemporary AI systems is that they lack a human-like understanding of causality \citep{Lake2016}. People utilize causality for classification and learning \citep{Rehder2001,Murphy2002}, explaining perceptual observations through hypothetical real-world generative processes. For instance, people group young and old trees together because they arise through the same causal process, while state-of-the-art computer visions systems based on convolutional neural networks \citep[ConvNets; ][]{LeCun1989} fail to see this type of similarity, even after training on a million natural images (Fig. \ref{fig:intro}A). Causality facilitates other types of everyday reasoning, such as predicting how a tree will grow, or how a tree could be trimmed to keep it healthy (Fig. \ref{fig:intro}B). These extrapolations require rich algorithmic abstractions and reasoning over multi-step latent processes, going beyond Bayesian networks \citep{Pearl2000} and the simple causal reasoning scenarios often studied in the lab \citep{Gopnik2004}, motivating the need for new technical tools with these rich causal capabilities.

To capture more causal and flexible types of learning, concept learning has been modeled as Bayesian program induction  \citep{LakeScience2015,Goodman2014,Stuhlmuller2010,Piantadosi2012,Ellis2015,Ellis2018}. Programs specify causal processes for generating examples of a category, utilizing high-level algorithmic units such as loops and recursion and allowing concepts to be used flexibly for multiple tasks. Bayesian inference is the engine for learning causal processes from examples, specifying the tradeoff between fitting the perceptual observations and favoring simpler generative programs. In previous work, Bayesian ``motor'' program induction has been used to learn new handwritten characters from a single example, leading to human-like patterns of generalization in various generative tasks \citep{LakeScience2015,LakeOmniglotProgress}. Beyond visual concept learning, program induction has also been applied to domains such as number word acquisition \citep{Piantadosi2012} and problem solving with toy trains \citep{Khemlani2013}.

This previous work covers interesting yet specialized cases: motor programs for generating characters or rearranging toy trains have unusually concrete and embodied semantics that are not representative of all programs. Moreover, program induction has steep computational demands, and traditional inference algorithms struggle to search the non-smooth and combinatorial program spaces. If the mind can induce genuine programs to represent concepts, what are the limits of this ability? Do people need explicit instruction regarding the underlying causal process -- as in practice writing handwritten letters -- or can people infer the causal process from just its outputs? Do mental concepts naturally include powerful computational techniques such as recursion?

Program induction over abstract recursive structures is not just a theoretical exercise. Recursion is central to language and thought \citep{Hauser2002,Corballis2014}, and many natural categories arise through recursive generative processes (Fig. \ref{fig:intro}C). Visual concepts such as trees, Romanesco broccoli, peacock plumage, ice drippings, rivers, sand dunes, fingerprints, wax drippings, clouds, etc. are natural fractals -- objects with characteristic patterns that appear at every scale \citep{Mandelbrot1983}. Breaking off a piece of Romanesco, chopping off the branch of a tree, or zooming in on a subset of a river delta each results in miniature versions of the original. This signature invites learners to search for simpler causal processes that can explain the visual complexity.

In this paper, we studied how people and machines learn abstract, recursive visual concepts from examples. The tasks were designed to explore the limits of the human ability to infer structured programs from examples -- in terms of the difficulty of the concepts, the amount of data provided (just one or a few examples), and the range of ways people can generalize (both classification and generation). While examining human learning, our tasks also present a new challenge for computational cognitive modeling and machine learning. We develop a Bayesian program learning (BPL) model that learns recursive program-based concepts from examples \citep{LakeScience2015}, providing an ideal observer analysis of the tasks as well as a framework for algorithmic-level modeling with resource limitations. We compare with multiple alternative computational approaches that do not operationalize concept learning as program induction, including deep neural networks, classic pattern recognition algorithms, and a lesioned BPL model without recursion.

\begin{figure}[t]
\ifthenelse{\boolean{arxiv}}
{\centering\includegraphics[width=3.4in]{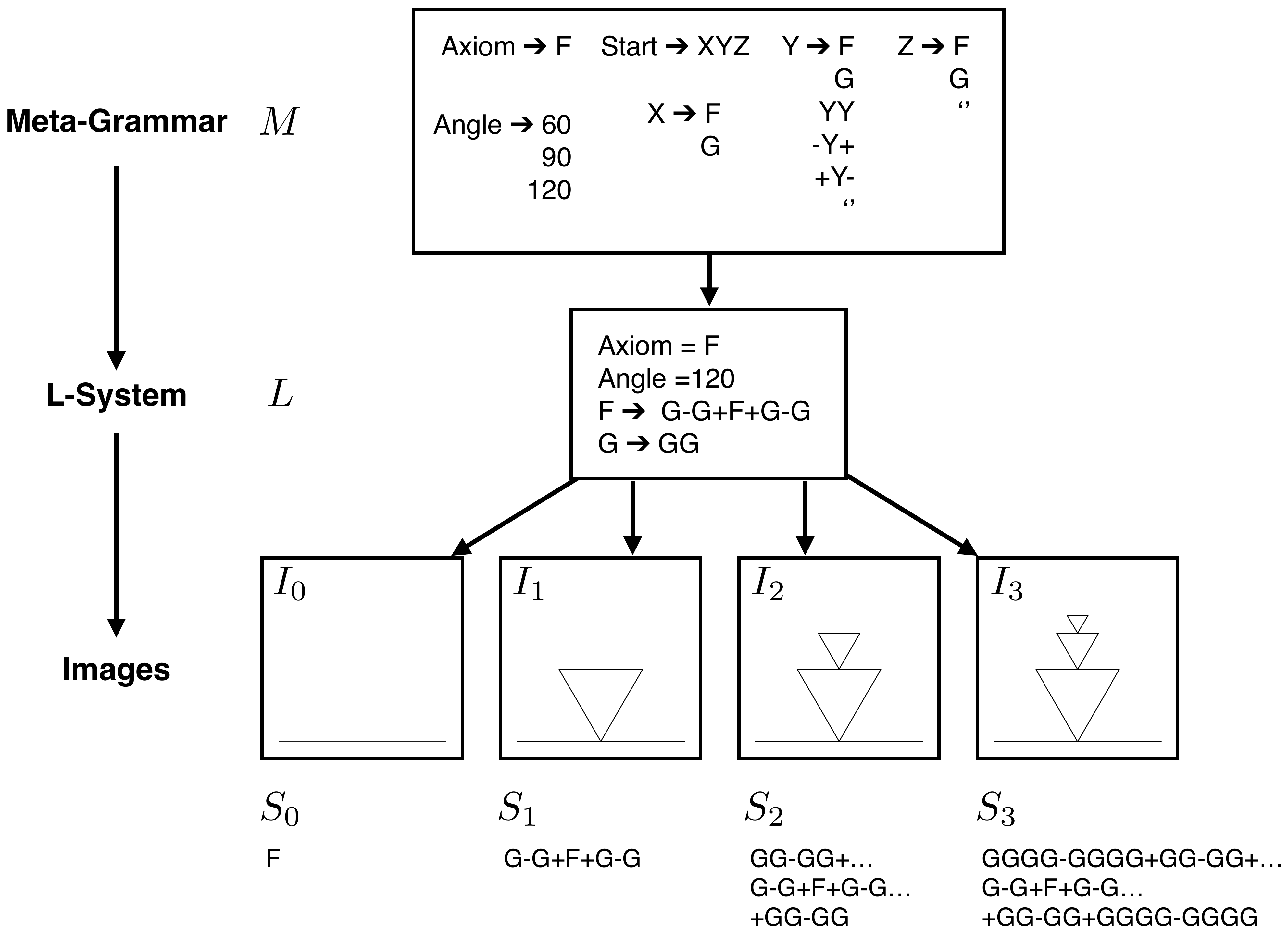}}
{\centering\includegraphics[width=5in]{figures/generative_model.pdf}}
\caption{A hierarchical generative model for recursive visual concepts.
A probabilistic context-free grammar ($M$) samples an L-system ($L$) which defines a type of image (a concept).
The L-system $L$ specifies an axiom, a turn angle, and re-write rules for the ``F'' and ``G'' symbols. 
Tokens of a concept have both a symbolic ($S_d$) and a visual ``Turtle graphics'' form ($I_d$), where $d$ indicates the depth of recursion.
In this example, the recursion operates as follows: the axiom ``F'' ($S_0$) is re-written to become ``G-G+F+G-G'' ($S_1$), which is rewritten to become $S_2$, and so on (the ``...'' indicates line breaks and are not symbols).
To transform $S_d$ into $I_d$, turtle starts at the bottom leftmost point of each figure with a rightward heading.}
\label{fig:generative_model}
\end{figure}

\section{Model}
We introduce a hierarchical Bayesian model for learning visual concepts from examples. During learning, the model receives a limited set of outputs (in this case, images) from an unknown program without the intermediate generative steps. The aim is to search the space of possible programs for those most likely to have generated those outputs, considering both sequential and recursive programs as candidate hypotheses. To construct the model, we first specify a language for visual concepts that is used both for generating the experimental stimuli and for the computational modeling. Second, we describe how to infer programs from their outputs through the hierarchical Bayesian framework.

\subsection*{A language for recursive visual concepts}
Lindenmayer systems (L-systems) provide a flexible language for recursive visual concepts, with applications to modeling cellular division, plant growth, and procedural graphics \citep{Lindenmayer1968a,Prusinkiewicz1990,Mech1996}. We use a class of L-systems that closely resemble context-free grammars, specifying a start symbol (axiom) and a set of symbol re-write rules. Each recursive application of the re-write rules produces a new string of symbols. Unlike context-free grammars that apply the re-write rules sequentially, L-systems apply all rules in parallel. As the rules are applied, each intermediate output is a different member of the category, which has both a symbolic ($S_d$) and visual interpretation ($I_d$), where $d$ indicates the depth of recursion.
An example L-system is shown in Fig. \ref{fig:generative_model}.

Building on prior work that studies figure perception as symbolic compression \citep{Leeuwenberg1969,Buffart1981}, the symbolic description ($S_d$) of an example is interpreted visually ($I_d$) using ``Turtle graphics'' (Fig. \ref{fig:generative_model}). Turtle graphics is a common environment for teaching programming, and it has been used in other program induction models as well \citep{Ellis2015}. The environment provides a virtual turtle that lives in 2D canvas with a location and orientation. The turtle can be controlled with simple instructions such as ``go straight'' (here, denoted by symbols ``F'' and ``G''), turn left (``-''), and turn right (``+''). As she moves, the turtle produces ink on the canvas. In this paper, turtle always moves a fixed number of steps, turns a fixed number of degrees, and produces ink while moving, although more complex control structures for the turtle are possible.

Equipped with these tools, our language builds in intermediate-level primitives and operators (e.g. line segments, angles, recursion, etc.), as opposed to a more basic (e.g. pixels or gabor filters) or more abstract levels of description (e.g. categories like ``trees"). A successful model, compared to alternatives formulated at other levels, would support this ``intermediate-level sketch'' as a cognitively natural level of description. The present language is certainly not complete: people undoubtedly utilize additional (or alternative) primitives and operations compared to the precise formalism introduced here -- a topic we take up in the general discussion. Nevertheless, our framework provides a reasonable starting point for studying how people learn program-like abstractions from examples, in terms of what level of description best explains learning and what conditions are required for generalization.

\subsection*{Bayesian program learning}
A computational model based on Bayesian Program Learning (BPL) model is used to infer an unknown program given just one or a small number of images produced by the program \citep[Fig. \ref{fig:generative_model}; ][]{LakeScience2015} \citep[see also, Probabilistic Language of Thought models; ][]{Piantadosi2011,Goodman2014,Overlan2016}.
The core of the BPL model is the hierarchical generative process shown in Fig. \ref{fig:generative_model}.
First, a meta-grammar $M$ samples a concept type, which is a L-system based program $L$.
To produce a token, the depth of recursion $d$ is either pre-specified or sampled from a uniform distribution (from 0 to 4).
The program $L$ is applied to its own output $d$ times, and the symbolic form $S_d$ is stochastically rendered as a binary image $I_d$.
The joint distribution of type $L$ and tokens $I_0,\dots,I_K$ is
\begin{equation} \label{joint_dist}
P(L,I_0,\dots,I_K) = P(L)\prod_{d=0}^{K} P(I_d|L).
\end{equation}
Concept learning becomes a problem for posterior inference, or reasoning about the distribution of programs given a set of images, either $P(L|I_0,\dots,I_K)$ (if the depths are pre-specified) or $P(L,j|I_j)$ (for a single image of unknown depth).

We now describe each term in Eq. \ref{joint_dist} to specify the full model.
The meta-grammar $M$ is a probabilistic context free grammar (PCFG) for generating $L$.
At each step, a PCFG re-write rule is chosen uniformly at random (from the applicable set) to define the prior on L-systems, $P(L)$. The random variables in $L$ are the turtle's turning angle and F-rule, the re-write rule for the ``F'' symbol. The PCFG generates a F-rule by beginning at ``Start'' and applying production rules until the string consists of only terminal symbols, $\{F,G,+,-,``\ "\}$ (Fig. \ref{fig:generative_model}). As a general class of grammars, L-systems can produce visual forms that are causally opaque, either because the form changes qualitatively at each iteration or because Turtle draws on-top of existing lines. Instead of working with this unrestricted class, we focus on L-systems that more closely resemble natural growth processes like those in Fig. \ref{fig:intro}, and for which people have a reasonable possibility of inferring the underlying growth process. At each iteration, these programs sprout symmetric growths from a subset of their straight line segments (``F'' symbols sprout and ``G'' symbols do not) that maintain the same global shape at each iteration (Fig. \ref{fig:generative_model}), with Turtle avoiding crossing previous paths. The details for generating $L$ from $M$ are provided in the Supplementary material.

Last, BPL requires an ink model to interface between the L-systems and the raw images they are meant to explain. To complete the forward model, an image $I_d$ is sampled from a stochastic process $P(I_d|L)$ that creates an image by computing the Turtle trajectory and sprinkling ink along the route.
First, $L$ is unrolled for $d$ iterations to produce a string of turtle symbols (Fig \ref{fig:generative_model}). Second, Turtle traces her trajectory, which is centered in the image frame and rescaled to have a common width. Third, a stochastic ink model transforms the real-valued trajectory into grayscale ink on pixels with discrete coordinates, using the approach developed in \citet{HintonNair2006} with the specific parameters used in \citet{LakeScience2015}. Each real-valued pixel defines the probability of producing a black pixel (rather than white) under an independent Bernoulli model.\footnote{Images presented to participants were rendered with standard Python graphics rather than the BPL ink model. The BPL ink model parameters were fit (via maximum likelihood) to the graphics using random turtle scribbles.}

To summarize, the BPL model specifies a grammar (PCFG) for generating another grammar-like program (L-system), and a process for expanding and rendering L-systems as raw images (via turtle graphics). The model can also solve the inverse problem: given an image, it can search for the underlying program (L-system) that is most likely to have generated it. Any given image is consistent with both recursive and non-recursive interpretations, and thus the model decides for itself which interpretation to take. An interpretation unrolled to $d\ge2$ iterations is a recursive generative process, while an interpretation with $d=1$ is a static, non-recursive generative process. Recovered programs can be run forward to generate further recursive iterations, generalizing beyond the input to perform a range of tasks.

To approximate Bayesian inference, we can draw posterior samples from $P(L|I_0,\dots,I_K)$ (pre-specified depth) or $P(L,j|I_j)$ (unknown depth) using Markov Chain Monte Carlo (MCMC) and a general inference scheme for grammatically structured hypothesis spaces. In short, a candidate L-system has a parse tree for generating it from the meta-grammar, and a proposal is made by regenerating a sub-tree from the meta-grammar to produce a new L-system  \citep{Goodman2008a,Piantadosi2012}. The algorithm was implemented in the LOTlib software package \citep{piantadosi2014lotlib}. This algorithm is effective for our problem, except that small changes to  hypotheses can sometimes produce very long symbolic forms ($S_d$) that require substantial computation to render as images. It is straightforward to rule out overly complex proposals without fully simulating their consequences, and thus the length of the sequences ($S_d$) was capped just above the length of the longest concept in the experiments. Hypotheses that exceeded this limit were decremented in recursive depth to $S_{d-1}$.

As an account of learning, it is important to state which components of BPL we see as critical and which are not. Our experiments examine whether people can learn recursive visual concepts from examples, and whether they engage with the underlying causal structure and its non-trivial algorithmic properties. Comparing models allows us to compare alternative levels of description, but it does not serve to identify the precise language and representational primitives that people use. In fact, in these experiments our instantiation of BPL has several advantages over people, reflecting its status as an ideal observer. The model starts with exactly the right internal programming language, allowing it to learn concepts in this family but not others. People do not have this internal language directly; instead, if they succeed on the tasks, their ``language of thought'' \citep{Fodor1975,Piantadosi2011,Goodman2014,Goodman2008} must be powerful enough to engage with the algorithmic properties of these stimuli, yet general enough to learn many other types of programs too. In sum, BPL provides an account of learning as inducing recursive generative programs, but we would not necessarily expect it to outperform alternative models that can also induce recursive generative programs, albeit with differing sets of primitives. The current experiments aim to distinguish models that engage with the algorithmic content of the stimuli from those that do not, such as the methods introduced below for computing visual similarity and for non-recursive program induction.

\subsection*{Alternative models}
We compare BPL with alternatives that do not utilize explicit program-like structures or abstractions, including three different approaches to measuring visual similarity based on generic features. The first uses the pre-trained visual features of a deep convolutional neural network \citep[ConvNet; ][]{Krizhevsky2012}, which is informed by extensive knowledge of objects and natural scenes instead of symbolic programs. The ConvNet is pre-trained on the large scale ImageNet object recognition challenge with 1.2 million natural images \citep{Russakovsky2014}, and similarity is measured as cosine distance in the top-most feature vector before the output layer. We also compare with Euclidean distance in pixel space and a classic metric in computer vision, the modified Hausdorff distance for comparing binary images \citep{Dubuisson1994}. These three metrics are not an exhaustive list of possibilities, but they represent standard techniques for measuring visual similarity that do not rely upon the type of explicit generative structure and algorithmic abstractions that BPL uses to generalize.

We also compare with a lesioned ``non-recursive BPL'' model restricted to depth $d\le1$, such that recursion cannot be used to explain the visual examples. The algorithm seeks to explain the (most mature) example with a complex sequence of Turtle commands, such that the most complex example in the experiment contains 470 symbols. To reduce the considerable search burden of finding these sequences, the model is provided with the ground truth generative sequence $S_j$ for the most mature example. Given the contour is modeled as a flat sequence, the likelihood of any new image $I_{j+1}$ is simply modeled as $P(I_{j+1}|S_j)$ without recursive expansion.

\section{Experiments}
Two experiments explored the human limits of inferring program-based concepts from examples. Participants were asked to learn new visual concepts from just one or a few examples of their outputs, and their ability to generalize was evaluated either through classifying new examples (experiment 1) or generating new examples (experiment 2). People, BPL, and the alternative models were compared on a set of tasks of varying difficulty, providing a comprehensive picture of the human ability and its boundaries. All of the experiments are available online,\footnote{\url{https://cims.nyu.edu/~brenden/supplemental/lrvc/vp-exp.html}} and the details are provided below.

\begin{figure}[!t]
\centering\includegraphics[width=3.4in]{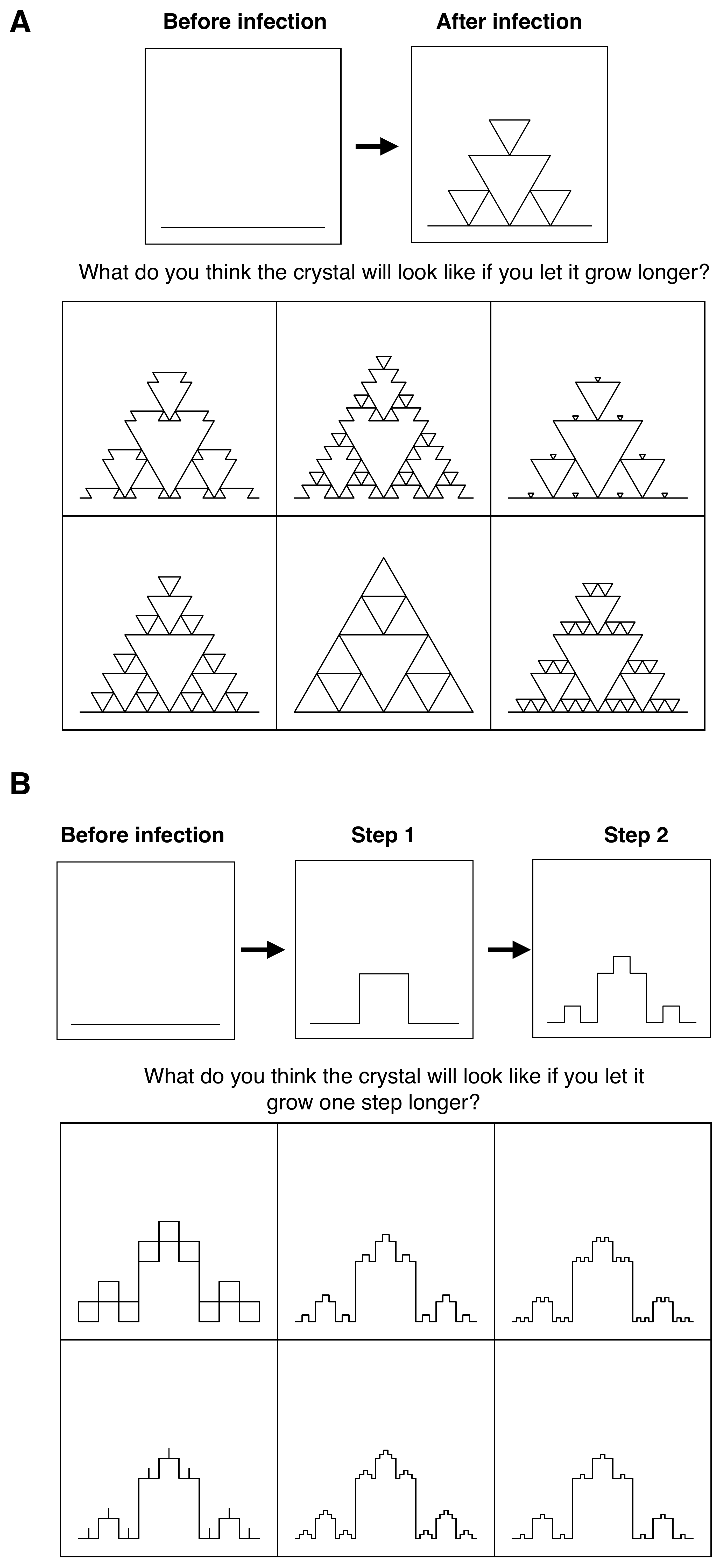}
\caption{Classifying a new example of a recursive visual concepts. Examples trials are shown for the block (A) vs. incremental condition (B). Answers: bottom-left (A) and top-middle (B).}
\label{fig:classif}
\end{figure}

\begin{figure}[!t]
\ifthenelse{\boolean{arxiv}}
{\centering\includegraphics[width=3.4in]{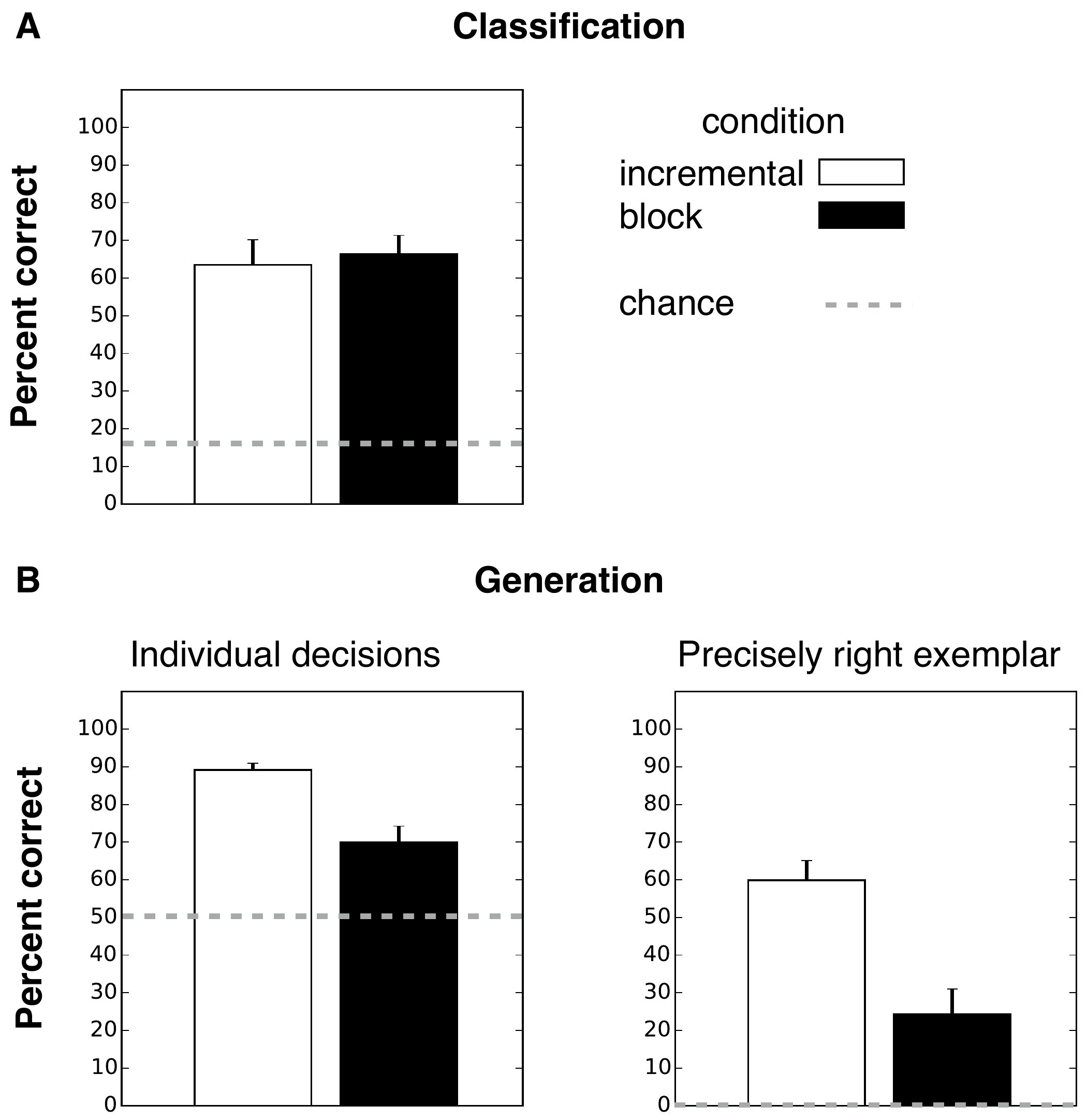}}
{\centering\includegraphics[width=4in]{figures/accuracy.pdf}}
\caption{Mean human performance on classification (A) and generation (B) tasks with recursive visual concepts. Accuracy for classification is based on a six-way choice. Accuracy for generation is measured on the basis of individual decisions (left) and whether exactly the right exemplar was produced (right). Error bars are $\pm$ SEM.}
\label{fig:acc}
\end{figure}

\subsection*{Experiment 1: Classification}
This experiment examines how people classify new examples of a recursive visual concept.

\subsubsection*{Methods}
Thirty participants in the United States were recruited on Amazon Mechanical Turk using psiTurk \citep{Psiturk}. Participants were paid \$2.50, and there was no special incentive for high accuracy. The experiment took an average of 11:53 minutes with a range from 3:52 to 36:51 minutes.

Participants were shown examples from 24 different recursive visual concepts and asked to classify new examples (Fig. \ref{fig:classif}).\footnote{One trial was removed after collecting the data because two visually identical distractors were mistakenly included.} Each of 24 trials introduce a separate concept, and participants made one classification judgment per trial. No feedback was provided to participants, in order to prevent supervised learning in the task. The instructions specified that each trial introduced ``a new type of alien crystal'' that had infected a surface and had been growing for some time. Participants were asked to predict what the crystal will look like as it continues to grow, and they were presented with a choice of six images. The stimuli were quite visually complex, and participants could magnify the details by rolling their mouse over a particular image area. After reading the instructions, participants were quizzed on their content and cycled back to re-read, until they got all of the comprehension questions correct \citep{Crump2013}.

Participants were assigned to one of two conditions that differ in the number of training examples: the ``incremental'' condition observed each step of growth ($d=\{0,1,2\}$; Fig. \ref{fig:classif}B), and the ``block'' condition observed only the final step of growth ($d=\{0,2\}$; Fig. \ref{fig:classif}A). These two conditions explore the boundary of the human ability in different ways, but we did not have strong a priori predictions regarding how this manipulation would influence behavior -- the ideal observer nature of the BPL model allows it to succeed in either condition. The incremental condition is an example of the Visual Recursive Task, and previous work has shown that both children and adults can perform the task successfully \citep{Martins2015a,Martins2014}. Our aim differs in that the incremental classification experiment is just the simplest of the evaluations we study. The more challenging ``block'' condition requires generalization from just a single static example of the concept (one-shot learning), and the generation task (experiment 2) probes richer forms of generalization beyond classification.

The 24 visual concepts were created by sampling L-systems from the BPL prior distribution. The provided examples were unrolled to depth $d=2$, and the task was to choose the natural extension (formally, the next iteration in depth $d=3$). The distractors in the forced-choice judgments were created by taking the example stimulus at $d=2$ (the ``After infection'' image in Fig. \ref{fig:classif}A or the ``Step 2'' image in Fig. \ref{fig:classif}B) and applying the expansion rules from a different L-system. The 24 concepts were sampled from the prior with a few additional constraints that standardized the concepts and eased cognitive penetrability: the fractal grows upwards, the turtle does not cross over her own path, and the F-rule expansion does not allow for two adjacent straight line symbols. Distractors were sampled without these constraints to ensure a sufficient variety of continuation types.

To perform the classification, BPL uses the last sample produced using MCMC to approximate the posterior predictive distribution, which is either $P(I_{3}|I_0,I_1,I_2)$ (incremental condition with known depth) or $P(I_{j+1}|I_j)$ (block condition with a single exemplar at unknown depth $j$). Each possible answer choice is scored according to this distribution, and the image option with highest posterior predictive probability is selected.

\begin{table}[t]
\centering
{\small
\begin{tabular}{lll}
 & classification & generation \\ \midrule
human behavior & 63.5\% & 59.9\% \\
BPL & 100\% & 100\% \\
\ -limited MCMC* & 64.5\% & 58.2\% \\
\ -without recursion & 30.4\% & 0\% \\
ConvNet & 4.4\% & 0\% \\
Euclidean & 30.4\% & 0\% \\
Hausdorff & 17.4\% & 0\% \\
Random & 16.7\% & 0\% \\
\bottomrule
\end{tabular}
}
\caption{Accuracy for humans and machines on the classification and generation tasks. Human performance is based on the incremental condition. Responses in the generation task were counted as correct only if participants produced exactly the right exemplar. (*) BPL with limited MCMC can achieve a range of different performance levels and was fit to match human performance.}
\label{table_model_perf}
\end{table}

\subsubsection*{Results}
Overall, participants extrapolated in ways consistent with the underlying program (Fig. \ref{fig:acc}A). The average accuracy across participants was 64.9\% ($SD=22.1$), which is significantly better than chance performance of 16.7\% ($t(29)=11.8$, $p < .001$). Performance was similar in both conditions, with 63.5\% correct in the incremental condition ($SD=25.0$, $n=15$) and 66.4\% in the block condition ($SD=18.5$, $n=15$; $t(28)=0.35$, $p > 0.5$). Average item accuracy was also correlated across the two conditions ($r=0.76$, $p<0.001$). Neither the pre-trained deep ConvNet nor Modified Hausdorff distance could classify the images better than chance (accuracy was 4.4\% and 17.4\%, respectively, choosing the test image that most closely matched the last training image). Evidently, people could intelligently reason about generative processes whether or not incremental steps were provided.

There was substantial variability in participant performance. Overall accuracy was correlated with time spent on the experiment ($r=0.61$, $p < .001$). In a post-experiment survey, participants who reported recognizing the stimuli as fractals performed better ($M=76.1\%$) than those who did not ($M=57.5\%$; $t(28)=2.4$, $p<0.05$). Importantly, even participants who did not recognize the stimuli as fractals performed above chance ($t(17)=7.84$, $p < .001$). Additionally, the degree of familiarity with fractals and whether or not a participant was a computer programmer did not significantly predict accuracy, which is noteworthy since recursion is an important technique in computer programming.

Performance of the computational models is summarized in Table \ref{table_model_perf}. Classification with BPL was simulated using MCMC for posterior inference. With enough samples, the model is a perfect classifier. To achieve high performance, the model depends crucially on its ability to learn recursive programs. In the block condition, the posterior mode program (aggregated across chains) was always recursive and correctly predicted the ground truth recursive depth, across all of the trials. In contrast, the non-recursive BPL model failed to classify new examples correctly, achieving only 30.4\% accuracy, performing at the level of the simple Euclidean distance metric.

As an ideal learner, BPL reaches perfect performance with enough samples, but people are not perfect classifiers. One way to analyze behavioral deviations is through failures of search \citep[or limited sampling; ][]{Liedera,Vul2014,Bramley2017}, with individual participants simulated as separate MCMC chains where some may fail to discover a suitable explanation of the data. With 15 simulated participants in each condition, 240 Metropolis-Hastings proposals for each chain provides the best match for  human-level accuracy, if decisions are made based on the last sample (64.5\% correct on average). This establishes only that BPL can achieve human-level accuracy or higher, while the four alternative models struggle to perform the classification task all together (Table \ref{table_model_perf}). However the limited MCMC model can be compared to behavior in other ways. Indeed, with this algorithmic limitation, BPL's predictions for which concepts are easier or harder to learn were moderately associated with human performance, in that the mean accuracy for the model and participants was correlated ($r=0.58$, $p < 0.01$, Supplementary material Fig. S1A). Unlike the alternatives, BPL can correctly classify the stimuli and predict which decisions are easier or harder for people.

Note, however, that it is possible that people used a heuristic to make classification decisions and that these heuristics only approximated BPL. In particular, choosing the image that contains a smaller (possibly rotated) version of the example image is closely related, but also distinct from, learning a recursive visual program. Although this heuristic does solve the classification task, it does not help to explain, as with BPL, why some concepts are easier to learn than others. Moreover, this heuristic does not specify how to generate new examples of the concept, which is the task that was evaluated next.

\subsection*{Experiment 2: Generation}
This experiment examines how people generate new examples of a recursive visual concept. Compared to classification experiment, this is a more difficult and unconstrained form of generalization that further explores the boundaries of the human ability.

\subsubsection*{Methods}
Thirty participants in the United States were recruited on Amazon Mechanical Turk using psiTurk. As before, participants were paid \$2.50 and there was no special incentive for high accuracy. The experiment took an average of 15:02 minutes with a range of 6:35 to 30:17 minutes.

\begin{figure}[!t]
\ifthenelse{\boolean{arxiv}}
{\centering\includegraphics[width=3.4in]{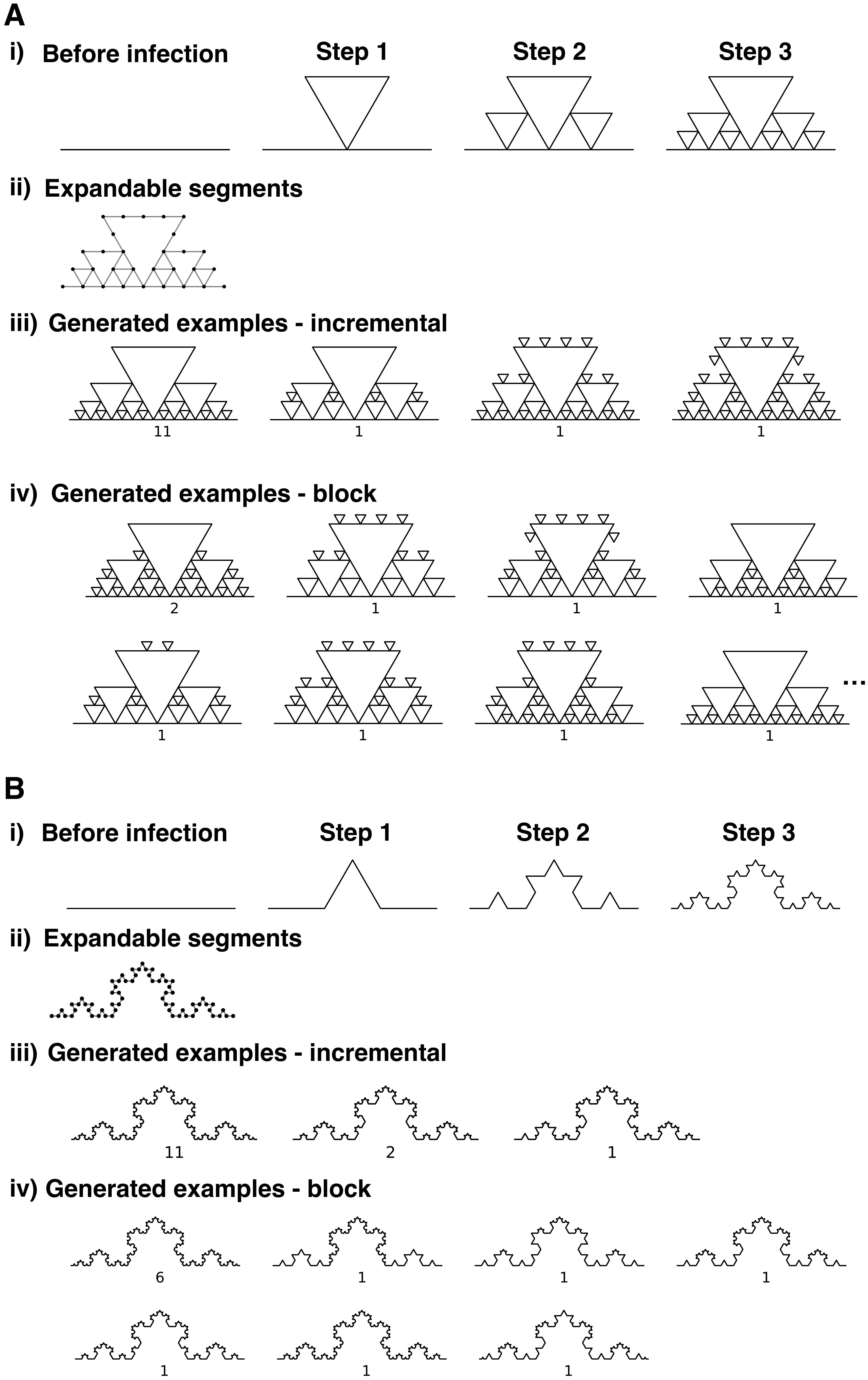}}
{\centering\includegraphics[width=4.25in]{figures/generation.pdf}}
\caption{Generating new examples of recursive visual concepts. Responses for individual participants are shown for two trials (A and B) with different concepts (i).  The incremental condition observed all three steps, while the block condition observed just step 0 and step 3.  An interactive display allowed participants to grow the figure by clicking on line segments of the example at step 3 (ii). Generated examples are shown in (iii) and (iv), and the number below each figure is the number of participants who generated it. The most frequently generated stimulus was correct in all cases except A-iv (not all responses shown for this group).}
\label{fig:gen}
\end{figure}

The procedures were adapted from experiment 1 and were the same except where noted. As before, participants were randomly assigned to either the incremental or block condition. There were 13 trials each with novel concepts, and example trials are shown in Fig. \ref{fig:gen}. In the incremental condition, participants saw novel concepts with three steps of growth (unrolled to depths $d=\{0,1,2,3\}$), and they were asked to predict just one additional step ($d=4$) beyond the most mature exemplar that was viewed. In the block condition, participants saw just the last, most mature step of growth ($d=\{0,3\}$) and were asked to demonstrate what the crystal will look like as it continues to grow ($d=4$). 

Participants used a custom web interface to generate the new example (Fig. \ref{fig:gen}ii). Clicking on a line segment toggles it from a deactivated state to an activated state (turning a ``G'' into an ``F'' in the symbolic language), or vice versa. Moving the mouse over a line segment highlights it in another color, and the color reveals how a click would affect the state. When highlighted green, clicking the segment sprouts a growth, activating it. When highlighted red, the segment is already activated and clicking it causes it to deactivate. Shortcut buttons allow participants to activate or deactivate all of the segments with a single click. Participants could interact with the display for as long as they needed.

Participants were informed that the order of their actions was not important, and only the final product mattered.
For some of the concepts in the previous experiment, the segments were too small for participants to effectively see and click; thus, this experiment used the 13 concepts from the classification experiment with the largest line segments. For 3 of the 13 concepts, some of the segments are redundant in that they create the same visual growth when activated as other segments (see the example in Fig. \ref{fig:gen}A-ii). For this reason, accuracy was scored according to the resulting visual form rather than the individual clicks, since different click patterns could result in the same visual pattern.

Four participants were excluded from the analysis. Two participants activated all of the growths for every single trial, and one participant did so for all but one trial. The data failed to record for one participant.

To generate a new example, BPL uses the last sample from MCMC to approximate the posterior predictive distribution, which is either $P(I_{4}|I_0,\dots,I_3)$ (incremental condition with known depth) or $P(I_{j+1}|I_j)$ (block condition with a single exemplar at unknown depth $j$). BPL makes a response using the same interactive display that participants used, allowing for $2^m$ possible responses given a display with $m$ expandable segment. To make a choice, the model visits each segment once in random order, and greedily decides whether or not to expand it in order to maximize the posterior predictive probability.

\subsubsection*{Results}
Participants generated exemplars that were highly structured and generally consistent with the underlying program, although there was substantial variability across participants (Fig. \ref{fig:gen}iii-iv).
Since each trial consists of many individual judgments (stimuli ranged between 22 and 125 segments), the accuracy of a trial was first computed by averaging across individual segment decisions, and then the accuracy of a participant was computed by averaging across trials.
Since growths tend to be sparse in the ground truth concepts, a baseline that deactivates all segments achieves 57.7\% correct.
Both groups performed significantly better than baseline (Fig. \ref{fig:acc}B): the incremental condition achieved 89.2\% correct ($SD=6.5$, $t(13)=17.4$, $p<0.001$) and the block condition achieved 70.0\% correct ($SD=14.2$, $t(11)=2.86$, $p < 0.05$).
The difference in means between groups was also significant ($t(24)=4.36$, $p < 0.001$).

Remarkably, participants were able to generate precisely the right example on a substantial number of trials, such that the example was only marked as correct if every individual segment was correct (Fig. \ref{fig:acc}B).
A random responder is effectively guaranteed to perform at 0\% correct, since even the simplest trial has over 4 million possible responses.
Alternatively, 3 of the 13 exemplars could be produced correctly by activating all of the segments in the interface (23.1\% correct).
Participants in the incremental condition produced precisely the right exemplar in 59.9\% of cases ($SD=18.9$), while participants in the block condition did so in 24.4\% of cases ($SD=21.9$; difference in means was significant, $t(24)=4.26$, $p < 0.001$).
Although both groups were far better than the random baseline, only the incremental group was significantly better than 23.1\% baseline on this conservative measure of accuracy ($t(13)=7.03$, $p<.001$). Thus, participants were accurate in both individual decisions and in aggregate, producing up to 125 decisions correctly to generate the right example. In the most difficult condition, participants produced exactly the right example only a quarter of the time, even if their accuracy was 70\% on individual decisions, suggesting people's inductive capabilities were nearing their limits.

BPL can also use the interactive interface to generate new examples. If the MCMC simulation is run for long enough, BPL achieves perfect performance, demonstrating that BPL can successfully produce new examples of recursive visual concepts. As with classification, we used short MCMC chains to simulate individual participants and modeled a response based on the last sample. The number of MCMC steps was fit to match human accuracy for generating exactly the right example, finding 160 steps matches the incremental group and 80 steps matches the block group. Unlike the classification task, MCMC alone did not predict which items are easier for participants, and model and participant accuracies were not significantly correlated ($r = 0.11$ and $r = -0.16$ for incremental and block groups, respectively; Supplementary material Fig. S2). Instead, properties of the response interface were the driving factors in item difficulty. Examples that can be generated with one or two well-placed actions were easier, while examples that required many actions were more difficult. For instance, the concept in Fig. \ref{fig:gen}B can be correctly extrapolated by activating all the segments with the appropriate shortcut button, while the concept in Fig. \ref{fig:gen}A requires eight individual activation actions. Assuming participants can begin from a fully activated or deactivated display, item accuracy is predicted by the number of required actions to produce the correct exemplar, with a correlation of $r=-0.66$ ($p < 0.05$) for the incremental condition and $r=-0.61$ ($p < 0.05$) for the block condition (Supplementary material Fig. S1B). This effect can be reproduced by the BPL model with response noise when acting using the response interface, but in our simulations this did not account for additional variance beyond the number of optimal actions.

Performance of the different computational models is compared in Table \ref{table_model_perf}. Of the range of models compared, BPL is the only computational model that learns recursive programs, and likewise it is the only model that successfully generates new examples of a novel concept. In contrast, the ConvNet, modified Hausdorff distance, Euclidean distance, and non-recursive BPL utterly fail at this task. For these algorithms, the best response is always to (incorrectly) create a new exemplar with zero growths activated, since it is maximally similar to the previous exemplar. Instead, recursive program-like representations provide an account of how a range of generalizations are possible from only the briefest exposure to a new concept.

\section{Discussion}

Compared to the best object recognition systems, people learn richer concepts from fewer examples.
Recent research in cognitive science, machine learning, and computer vision has begun to model learning as a form of program induction \citep{Zhu2006,Savova2008,Savova2009,Stuhlmuller2010,Piantadosi2011,Piantadosi2012,Khemlani2013,Goodman2014,LakeScience2015,Ellis2015,yildirim2015learning,rothe2017question,amalric2017language,Ellis2018,depeweg2018solving}, yet there are steep computational obstacles to building a general purpose program learner.
It is unclear how the mind could learn genuine programs in a general way.

Here, we explored the boundaries of the human ability to learn programs from examples.
We probed several key dimensions in a concept learning task: the difficulty of the concepts, the number of examples, and the format of generalization.
We found that people could both classify and generate new examples in ways consistent with a Bayesian program learning model, even though the model was provided with substantial knowledge about the structure of the concept space.
In a classification task that fools the best object recognition algorithms, participants responded with high accuracy and in accordance with the underlying recursive structure, regardless of whether they saw just one or two examples (block versus incremental condition).
In a more challenging generation task, people constructed new examples that were consistent with the underlying programs.
For generation, additional examples provided a boost in accuracy (three examples versus one example), while the one-shot case proved taxing and approached the boundary of people's inductive abilities.

People's success contrasts with the performance of pre-trained object recognition systems (ConvNets) and other pattern recognition techniques that do not explicitly represent causal processes and algorithmic abstractions like recursion.
Although feature-based approaches have proven effective in machine learning and computer vision, causal modeling and program induction hold promise for additional advances.
Causal representations are central to human perceptual and conceptual abilities \citep{Murphy1985,Gelman2003,Leyton2003,Rehder2001,Bever2010}, and they can also form the basis for high performance classification and prediction algorithms \citep{LakeScience2015,Pearl2019}.
Causal models can help explain wide variations in appearance without the need for extensive training data-- for instance, highlighting the commonalities between young and old trees of the same species, despite dramatic differences in their superficial features (Fig \ref{fig:intro}A).
Causal knowledge can also inform other types of everyday conceptual judgments (Fig \ref{fig:intro}B):
Is this tree growing too close to my house?
What will it look like next summer as it continues to grow?

There are several straightforward and meaningful extensions of the representation language studied here. Although the current concepts have some stochastic properties, including the depth of recursion and the stochastic renderer, they are more deterministic than their natural analogs (Fig. \ref{fig:intro}C). Our concepts grow in an orderly sequence of recursive steps, while natural growth is far more stochastic and only imperfectly scale-invariant. In future work, we intend to explore stochastic L-systems to address some of these challenges \citep{Prusinkiewicz1990}. Both failures to grow and spontaneous growths could be modeled with noise that flips ``F'' symbols (growth) to become ``G'' symbols (non-growth) and vice versa with a small probability, before applying the re-write rules at each recursive step. Additionally, context-sensitive L-systems could be used  as a more powerful language for generative processes \citep{Prusinkiewicz1990}, necessary for representing environmentally-sensitive growth processes like a tree that grows around an obstacle. Finally, people likely have primitives for simple shapes like triangles, squares, and rectangles, while the current BPL implementation can produce these shapes but does not represent them differently than other contours. Providing models with a richer set of primitives could further help close the gap between human and machine performance. All of these extensions are compatible with our framework, and are necessary components of an account of concept learning as program induction.

Our tasks pose a new challenge problem for machine learning. Although one could explore different pre-training regimens for deep ConvNets, which could improve their performance on our tasks, an alternative is to explore learning generative, program-like representations with neural networks. There has been significant recent interest in ``differentiable programming,'' or using neural networks to learn simple types of programs from examples, including sorting \citep{Graves2014}, arithmetic \citep{Weston2015}, finding shortest paths in a graph \citep{Graves2016}, and learning compositional rules \citep{LakeMeta2019}. Developing alternative models through differentiable programming could further refine the account presented here, establishing which aspects of the representational language are essential and which are not. Domain general program induction remains an important computational and scientific goal, with potential to deepen our understanding of how people learn such rich concepts, from such little data, across such a wide range of domains.

Overall, our results suggest that the best program learning techniques will likely need to include explicit -- or easily formulated -- high-level computational abstractions. As demonstrated here, people can learn visual concepts with rich notions of recursion, growth, and graphical rendering from just one or a few examples. Computational approaches must similarly engage rich algorithmic content to achieve human-level concept learning.

\subsection*{Acknowledgments}
We gratefully acknowledge support from the Moore-Sloan Data Science Environment. We thank Philip Johnson-Laird and Sangeet Khemlani for helpful comments and suggestions, and Neil Bramley for providing comments on a preliminary draft. 

\bibliographystyle{apacite}
\bibliography{library_clean}

\ifthenelse{\boolean{arxiv}}
{
  \clearpage
  

\section*{Supplementary material}

# Supplementary material:

# People infer recursive visual concepts from just a few examples

**Generating L-systems from the meta-grammar**

The L-system axiom is always "F". The angle is sampled uniformly from $\{60, 90, 120\}$. The constrained F-rule is generated as follows. The start symbol produces three non-terminals: "X" (prefix), "Y" (body), and "Z" (middle). After each non-terminal grounds out as a string of terminals, the right-hand-side of the F-rule is defined by the resulting string concatenation $X||Y||Z||\text{reverse}(Y)||X$, where $||$ concatenates and $\text{reverse}(\cdot)$ reorders string. For instance, if $X = G$, $Y = -G+$, and $Z = F$, the rule F ← G-G+F+G-G is produced. The G-rule is defined deterministically given the F-rule, so that all line segments ("F" and "G") grow at the same rate at each iteration.

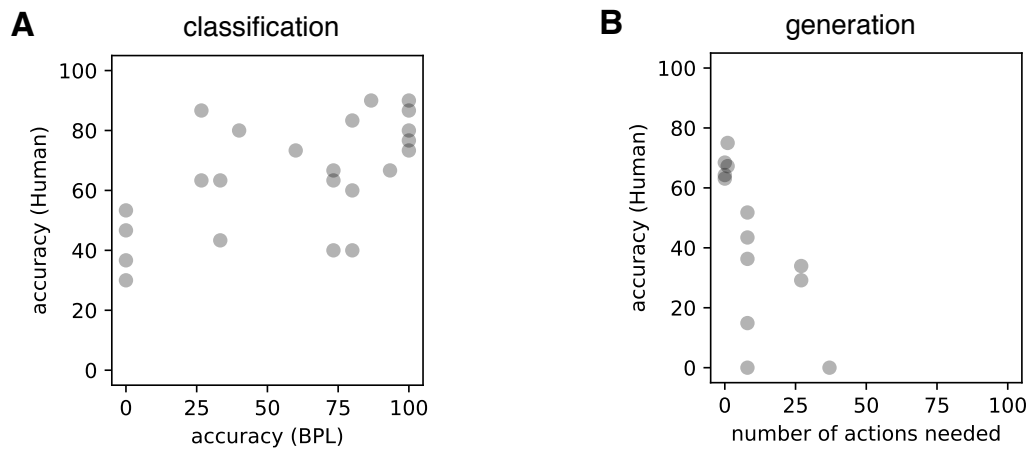

Figure S1: Item analysis for classification (A) and generation (B) tasks. Each point represents a different recursive visual concept. Classification accuracy is measured as performance in the six-way classification task, and generation accuracy is measured by marking exemplars as correct only if every individual segment is correct. The behavioral data is aggregated across incremental and block conditions. Item difficulty is best predicted by BPL with limited MCMC for classification (A) and the number of required actions to produce the correct exemplar for generation (B).

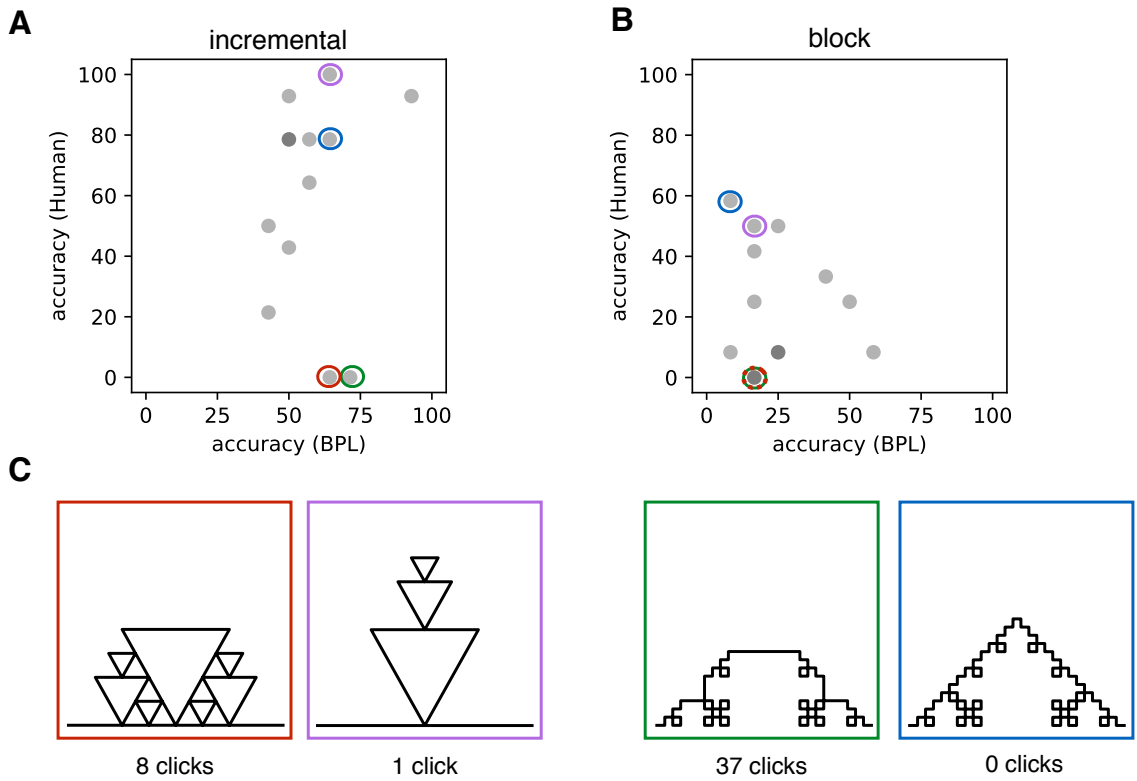

Figure S2: Item analysis for generation task. Each point represents one of the 13 different recursive visual concepts, both for the incremental group (A) and the block group (B). Accuracy is measured by marking exemplars as correct only if every individual segment is correct. Individual concepts (C) are shown expanded to depth 3, while the task was to generate a new example at depth 4. The concepts highlighted in red and purple have similar complexity for BPL with limited MCMC (same re-write rule but with different placements of 'F' and 'G' symbols), while the red one is far more difficult for participants to interpret. Likewise for the concepts highlighted in green and blue. However, the number of actions required to produce the correct exemplar (listed below each concept) does predict the difficulty of the item for participants, where exemplars that require fewer clicks are easier (see also Fig. S1B).

15


}
{}
\end{document}